\documentclass{article}
\usepackage{spconf,amsmath,graphicx}
\usepackage{threeparttable}
\usepackage{amsfonts}
\usepackage{xr}
\usepackage{multirow}
\usepackage{cite}
\usepackage{svg}
\usepackage{url}
\usepackage{xcolor}

\usepackage{hyperref}

\usepackage{graphicx}
\usepackage{textcomp}
\usepackage{xcolor}
\usepackage{comment}
\usepackage{blindtext}
\usepackage{siunitx}
\usepackage{graphicx}
\usepackage{subcaption}
\usepackage{balance}
\usepackage{bm}
\usepackage{multirow}
\usepackage{booktabs}
\usepackage{pifont}
\newcommand{\xmark}{\ding{55}}%
\usepackage[linesnumbered,ruled,vlined]{algorithm2e}

\title{Contrastive Deep Encoding Enables Uncertainty-aware Machine-learning-assisted Histopathology}

\name{\normalfont \normalsize \begin{tabular}{c}Nirhoshan Sivaroopan$^{1, \dagger}$,  Chamuditha Jayanga$^{1, \dagger}$, Chalani Ekanayake$^{1, \dagger}$, Hasindri Watawana$^{1, \dagger}$, \\ Jathurshan Pradeepkumar$^{2}$, Mithunjha Anandakumar$^{2}$, 
Ranga Rodrigo$^{1}$,  Chamira U. S. Edussooriya$^{1}$,\\ \& Dushan N. Wadduwage$^{2,*}$ \thanks{$^{\dagger}$ These authors contributed equally to the work.}\end{tabular}}
  
\address{\normalsize $^{1}$ Dept. of Electronic and Telecommunication Engineering, University of Moratuwa, Sri Lanka \\
\normalsize $^{2}$ Center for Advanced Imaging, Harvard University, Cambridge, MA, 02138 \\
\normalsize $^{*}$ \href{mailto:wadduwage@fas.harvard.edu}{wadduwage@fas.harvard.edu}}

\begin{document}

%
\maketitle
\begin{abstract}
Deep neural network models can learn clinically relevant features from millions of histopathology images. However generating high-quality annotations to train such models for each hospital, each cancer type, and each diagnostic task is prohibitively laborious. On the other hand, terabytes of training data ---while lacking reliable annotations--- are readily available in the public domain in some cases. In this work, we explore how these large datasets can be consciously utilized to pre-train deep networks to encode informative representations. We then fine-tune our pre-trained models on a fraction of annotated training data to perform specific downstream tasks. We show that our approach can reach the state-of-the-art (SOTA) for patch-level classification with only 1-10\% randomly selected annotations compared to other SOTA approaches. Moreover, we propose an uncertainty-aware loss function, to quantify the model confidence during inference. Quantified uncertainty helps experts select the best instances to label for further training. Our uncertainty-aware labeling reaches the SOTA with significantly fewer annotations compared to random labeling. Last, we demonstrate how our pre-trained encoders can surpass current SOTA for whole-slide image classification with weak supervision. Our work lays the foundation for data and task-agnostic pre-trained deep networks with quantified uncertainty.

\end{abstract}


\vspace{-2mm}
\section{Introduction}
\label{sec:intro}
\vspace{-1mm}
Computer-assisted histopathological diagnostics using deep learning is an emerging field. Especially for cancer, deep models have demonstrated their potential in the clinic by effectively streamlining labor-intensive and error-prone image-based detection procedures, thereby complementing pathologists in the diagnostic process. For instance, in the detection of breast cancer metastases, a prevalent form of cancer, the early work by Wang et al. \cite{wang2016deep} has made significant strides in reducing human error rates by integrating deep learning models with human pathologists' assessments. Similarly, during the last decade, deep learning has demonstrated its capability as a promising tool in digital pathology, with numerous studies highlighting its diverse clinical applications. These include clinical diagnosis \cite{masud2021machine,jiang2020emerging,wang2017machine,lu2020deep}, prognosis/survival prediction \cite{li2020machine,wulczyn2020deep,zhu2016deep,kather2019predicting}, treatment response forecasting \cite{byra2020early,lu2021deep}, and the identification of regions of interest (RoIs) that exhibit substantial diagnostic value \cite{liu2021review,guo2019deep,liu2023deep}. Thus, machine-learning-driven digital pathology can potentially enhance multiple facets of the clinical process including accuracy, analysis speed, and reproducibility \cite{rashidi2019artificial,sakamoto2020narrative,harrison2021introduction}.

Despite notable achievements of deep learning in histopathology, several challenges persist \cite{tizhoosh2018artificial}. First, unlike conventional images where the target object occupies a substantial portion of the image, cancer-positive histopathology images typically feature small regions of positive activations over a large area of normal tissue. One should therefore image many slides of biopsy specimens to collect a sufficiently large training dataset. Some cancer types are rare and access to such samples is limited. Furthermore, patient data cannot be readily shared due to privacy concerns. This lack of access to training data discourages community from developing better models. Second, to train a fully supervised model, images should be finely annotated by expert pathologists identifying visually intricate patterns critical for accurate diagnosis. Such careful annotations are time-consuming. Return on expert time invested in annotating is also not guaranteed in terms of the final performance of the model. This uncertainty discourages expert pathologists from spending time annotating large datasets to a standard of accuracy needed to train a reliable deep model in a fully supervised fashion. Third, most deep-learning models for computational pathology are not interpretable. The end user is not made aware of the uncertainty in model predictions. This lack of transparency makes it challenging to integrate these models into the existing clinical decision process. In this work we attempt to tackle all three of these challenges by: using publicly available large datasets; self-supervised pre-training followed by supervised fine-tuning with only a few annotations; and explicitly quantifying the level of uncertainty.

To this end, based on the seminal SimCLRv2 self-supervised framework \cite{chen2020big}, we introduce an uncertainty-aware learning method for digital pathology. Our work results in three significant advancements over the state-of-the-art (SOTA). First, we overcome annotation limitations by training an accurate model with minimal labeled training samples. Second, we establish our model as a versatile framework capable of adapting to various clinical tasks. Third, we quantify the uncertainty in our model predictions, empowering pathologists to make more informed decisions in the context of model confidence. With these contributions, our approach outperformed SOTA in both patch and slide-level predictions on multiple benchmark datasets while uniquely quantifying risks associated with uncertain model predictions.  

\section{Related Work}
\vspace{-1mm}
\subsection{Deep learning for digital pathology} 
\vspace{-2.5mm}
Some seminal work that laid foundation for machine-learning-assisted histopathology include: \cite{beck2011systematic} that applied logistic regression to investigate the correlation between histology of the surrounding connective tissue of tumor cells and prognosis in breast cancer; \cite{yu2016predicting} that studied survival prediction for lung cancers using regularized machine learning and automatic feature extraction; and \cite{xu2017large} that visually examined individual node responses in the last hidden layer of a convolutional neural network to uncover biologically insightful interpretable characteristics in histopathology images.
\vspace{-4mm}

\subsection{Self-supervised representation learning}
\vspace{-2.5mm}
Self-supervised learning (SSL)\cite{10.5555/3454287.3455679,Misra2019SelfSupervisedLO,what,NEURIPS2020_70feb62b,byol,sim,obow,barlow} has proven successful in computer vision to pre-train large models with unlabelled data. SimCLR\cite{chen2020simple}, a contrastive learning framework, is an example of this approach to learn visual representations of images. SimCLRv2 \cite{chen2020big} enhanced SimCLR in multiple ways: utilizing larger backbone networks during pre-training, increasing the capacity of the projection head, incorporating a memory mechanism from MoCo \cite{he2019momentum}, and leveraging knowledge distillation \cite{hinton2015distilling} using unlabeled examples. Masked auto-encoding (MAE) \cite{mae} is another SSL approach that reconstructs signals from partial observations. It consists of an encoder that maps a partially masked image to a latent representation and a decoder that reconstructs the original image from it. By masking $~75\%$ of the input patches of the image, MAE enabled scalable training of large models using a reconstruction loss on masked patches. The efficacy of MAE has been demonstrated in various downstream tasks, including image classification. 

Both SimCLR and MAE allow learning of rich representations from large amounts of unlabeled data and have also been effectively used in digital pathology. \cite{SelfSupervisedHisto} adapted SimCLR for digital pathology. \cite{luo2022self} used MAE in digital pathology and introduced a self-distillation loss on top of the masked patch reconstruction loss from original MAE. Nevertheless, more advanced SimCLRv2 has not been adopted for digital pathology. Moreover, none of these SSL models have been investigated in relation to the uncertainty of their predictions.

\vspace{-3mm}

\subsection{Uncertainty quantification}
\vspace{-2.5mm}
Estimating uncertainty of deep learning models is an active area of reaserch in machine learning. A straightforward approach is to use the model to output a distribution of  predictions  for each input rather than a single prediction. The variability of the predicted distribution can then be used to quantify uncertainty. Monte Carlo dropout method is one such approach, where each input is forward passed multiple times through the model (during both training and inference), while randomly dropping out different network nodes. Here, the standard deviation of the generated distribution of outputs is an estimator for uncertainty \cite{wang2013fast,gal2016dropout}. Deep ensembles is another method to quantify uncertainty. It generates a distribution of predictions by forwarding each input through a set of deep learning models that are trained with different initializations and hyperparameters \cite{lakshminarayanan2017simple,wenzel2020hyperparameter}. Entropy of the model output is then used to estimate the uncertainty. Test-time-augmentation \cite{dolezal2022uncertainty} is another uncertainty quantification method. It applies random transformations to each image and obtains predictions for each. Prediction variance among the perturbed versions of the same image is used as the uncertainty estimation. 

Despite the critical clinical importance, most deep learning frameworks for histopathology do not quantify the confidence of model predictions. A notable exception is \cite{dolezal2022uncertainty} that quantified histologic ambiguity of images using an uncertainty score. The uncertainty here is estimated during inference by sending each image tile through 30 forward passes in a dropout-enabled network, resulting in a distribution of predictions. The standard deviation of predictions for a single image patch represents patch-level uncertainty. These uncertainty quantification methods rely on a parameter with inherent ambiguity, rather than providing a precise mathematical quantification of the uncertainty level. 

In our work we augmented SSL frameworks to explicitly include uncertainty estimation. We adopted a Bayesian approach first proposed by \cite{sensoy2018evidential} using the theory of evidence . \cite{sensoy2018evidential} compared this Bayesian approach with others to show its utility for out-domain training.  As we focus on task-agnostic out-domain learning from public datasets, we leverage this method in our work.

\section{Results}
\label{sec:experiments}
\vspace{-1mm}
\begin{figure*}[h]
    \centering
    \includegraphics[width=\linewidth]{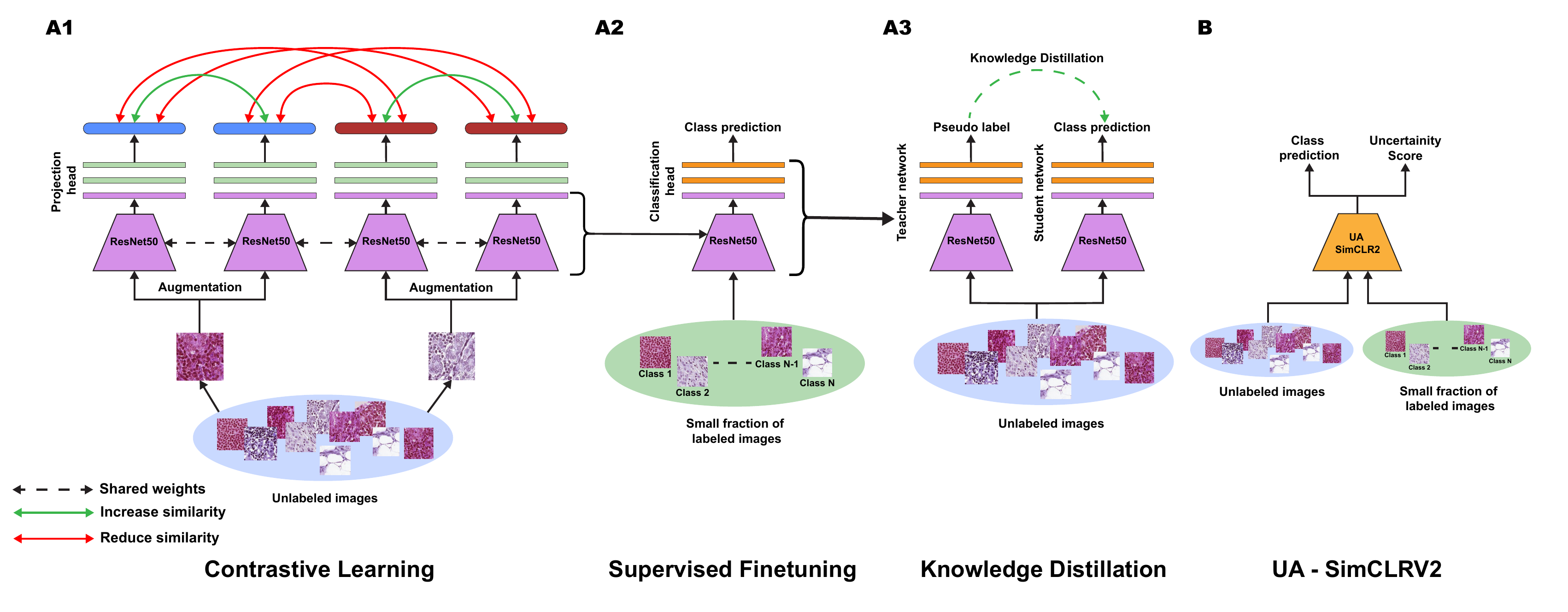}
    \vspace{-6mm}
    \caption{\small The SimCLRv2 framework. (A1) The pre-training step. Contrastive Learning is used to pre-train a deep neural encoder using a large set of unlabelled images. (A2) The supervised fine-tuning step. A classification head is added to the pre-trained encoder.  The model is then fine-tuned using a small fraction of labeled images. (A3) The knowledge distillation step. The model from 'A2' is used as a teacher network to generate pseudo labels for all unlabeled training images. Then the pseudo labels are used to train a student network (with the same architecture). (B) The proposed Uncertainty-aware(UA) SimCLRv2 model with an additional output for the uncertainty score.}
    \label{fig:fig1}
\end{figure*}

\subsection{SimCLRv2 and uncertainty-aware (UA-) SimCLRv2}
\vspace{-3mm}
We experimented with two state-of-the-art self supervised pre-training approaches on two publicly available patch-level datasets. We tested masked auto-encoding (MAE)\cite{mae} using a transformer backbone and contrastive learning using SimCLR frameworks (v1\cite{chen2020simple} and v2\cite{NEURIPS2020_fcbc95cc}). Out of these SimCLR-v2 performed best and hence was selected for this work. The training procedure of SimCLRv2 involves three stages (see Fig.~\ref{fig:fig1}.A1-3). First, a ResNet backbone undergoes self-supervised pre-training, where it learns representations through a stack of projection layers (see Fig.~\ref{fig:fig1}.A1). This process utilizes a contrastive loss function to compare the embeddings generated by the projected head. Second, a classification head is appended to the pre-trained encoder, and the model is fine-tuned fully-supervised using a small fraction of annotated patches (Ref. Fig.~\ref{fig:fig1}.A2). Last, the learned classifier, i.e. the teacher model, transfers its knowledge to an architecturally identical student model through a process known as knowledge distillation (see Fig.~\ref{fig:fig1}.A3). Here the teacher's pseudo labels (provisional labels assigned to unlabeled data based on model predictions) on the entire large unlabelled dataset are used as ground-truths for the student model. Note that the models use annotations only in the fine-tuning stage (see Fig.~\ref{fig:fig1}.A2). 

\begin{figure*}[htbp]
    \centering
    \includegraphics[width=0.7\linewidth]{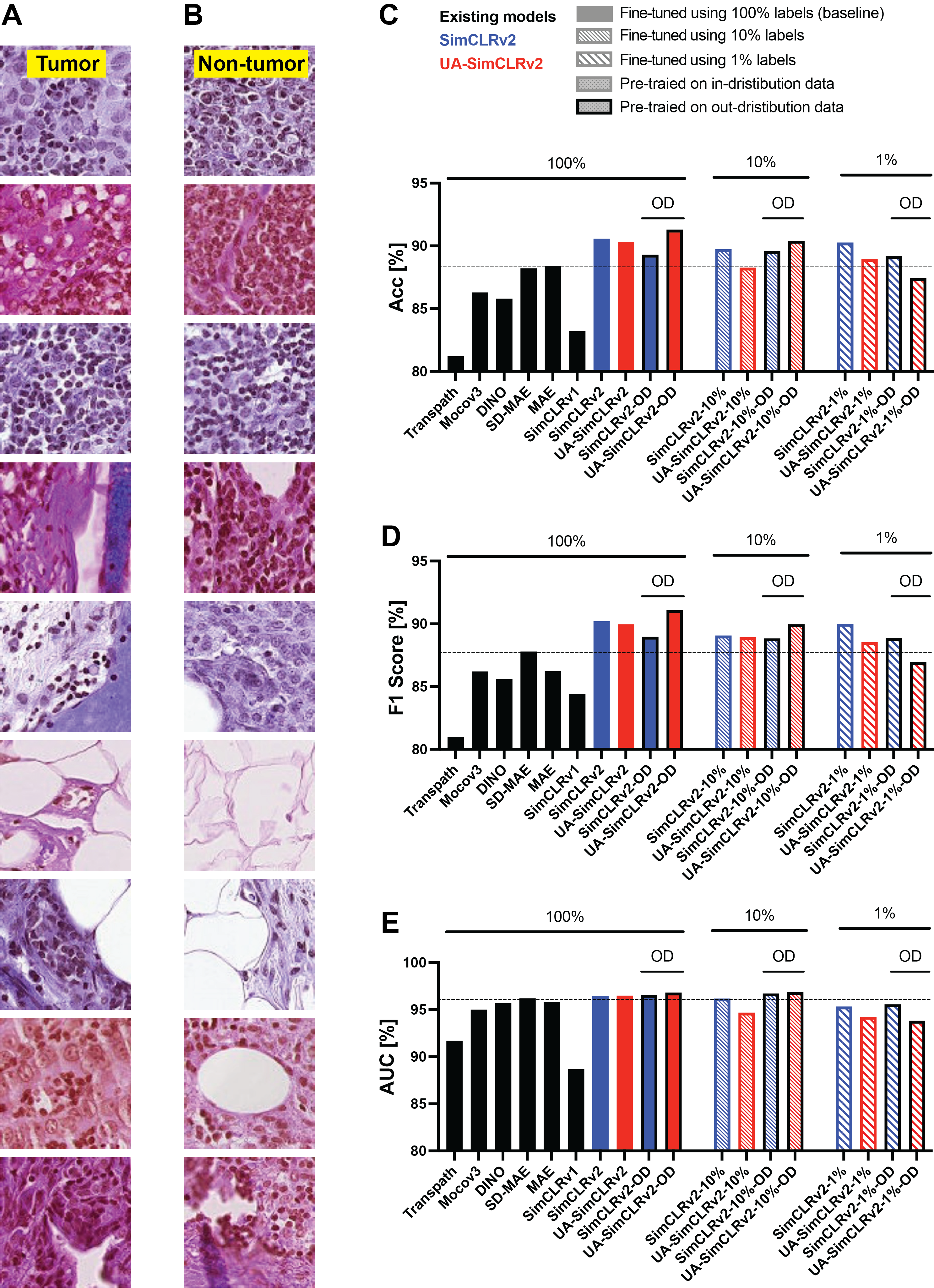}
    \caption{\small Binary classification results for PCam dataset for an ensemble of models. (A) Representative images from the tumor class. (B) Representative images from the non-tumor class. (C) Classification accuracy for competing models (shown in black) vs. the proposed models: SimCLRv2 and UA-SimCLRv2 (i.e. Uncertainty Aware SimCLRv2). The proposed models were first trained on 100\% of annotations to generate the baselines. Models were pretrained using in-distribution data (i.e. training data from NCT100k) as well as out-of-distribution data (i.e. training data from PCam). Then the same experiments were repeated with 10\% of annotations, and 1\% of annotations. (D) F1-scores for the same experiments in ‘C’. (E) Area under the curve (AUC) for the same classification experiments in ‘C’. }
    \label{fig:fig2}
\end{figure*}
While the SimCLRv2 model with a ResNet-50 backbone demonstrates superior performance in various clinical tasks (see our results in the next section), it lacks interpretability. Medical artifical intelligence (AI) applications, however,  require transparent and explainable results due to the potential risks associated with predictions. To address this issue, we propose a modified version of SimCLRv2 called Uncertainty-aware SimCLRv2 (UA-SimCLRv2) shown in Fig.~\ref{fig:fig1}.B. UA-SimCLRv2 follows a similar pre-training process as SimCLRv2; however, it undergoes fine-tuning and distillation using MSE loss in the Bayesian view. Notably, UA-SimCLRv2 not only provides class labels but also provides an associated uncertainty score for each patch prediction. We discuss UA-SimCLRv2 in detail in the methods section~\ref{sec:Methodology}.

In the next section, we present SimCLRv2 and UA-SimCLRv2 results for patch-level classification tasks in the digital pathology domain. 

\subsection{Patch classification results using SimCLRv2 and UA-SimCLRv2}
\vspace{-3mm}
Histopathology studies analyze data at multiple levels of hierarchy including patch-level, slide-level, and patient-level. We first employ SimCLRv2 and UA-SimCLRv2 as patch-level classifiers (later we adapt patch-level trained models to slide-level). In our study, we utilized two datasets: PatchCamelyon (PCam) and NCT-CRC-HE-100k (NCT100k). The experimental setup for evaluating model performances on these datasets included two variables: (1) the percentage of annotations used for fine-tuning from the training set, and (2) whether pre-training was performed on in-distribution data or out-distribution data. For example, in PCam experiments, out-distribution-pre-training means that pre-training was done on the NCT-CRC-HE-100k dataset (and vice versa in NCT-CRC-HE-100k experiments).\newline 

\vspace{-3mm}
\noindent \textbf{Binary classification results on PCAM.} The PCam dataset \cite{Veeling2018-qh} comprised a collection of 327,680 patches extracted from histopathology scans of lymph node sections in the CAMELYON16 dataset \cite{10.1001/jama.2017.14585}. Each patch had a corresponding binary label indicating the presence of metastatic tissue. According to \cite{Veeling2018-qh}, a positive patch contained at least one pixel of tumor tissue in its center 32x32 pixel region. Patches were 96x96 pixels in size, but for consistency across experiments, we resized all patches to 224x224 pixels (see the methods section for more details). We divided the dataset into training (75\%), validation (12.5\%), and test (12.5\%) splits, ensuring a balanced distribution of positive and negative examples with no overlap between the splits. 

Fig.~\ref{fig:fig2} and Table~\ref{table: pcam2} show the accuracy, F1 score, and AUC score (Area Under the ROC Curve) for the  PCam binary classification. To establish a baseline, we first fine-tuned our models with all training labels (i.e. the 100\% setting). Here, our models outperformed the state-of-the-art (SOTA) approach, i.e., MAE \cite{mae}. Notably, out-distribution-pre-trained UA-SimCLRv2 performed best with 2.89\% increase in accuracy, 2.74\% in F1 score and 0.77\% in AUC score compared to the SOTA. 

Next, we fine-tuned our models on 10\%  training labels. 10\%-fine-tuned models performed slightly worse than the 100\% baseline. Nevertheless, the 10\%-fine-tuned SimCLRv2 and  UA-SimCLRv2 still performed on par with or better than the SOTA (see Fig.~\ref{fig:fig2} and Table~\ref{table: pcam2}). Last, we fine-tuned our models on 1\% training labels. Interestingly the SimCLRv2 and UA-SimCLRv2 models still performed comparable to the SOTA (see the 1\% setting on Fig.~\ref{fig:fig2} and Table~\ref{table: pcam2}). However, at the 1\% setting UA-SimCLRv2 consistently underperformed compared to SimCLRv2, perhaps due to the limited evidence available for uncertainty awareness (see method section~\ref{sec:Methodology} for more details). \newline



\begin{table*}[]
    \centering
    \small{
    \caption{\small Binary classification results for PCam dataset for an ensemble of models under 100\%, 10\%, and 1\% labels settings. For selected cases, MAE, SimCLRv1, and SimCLRv2 models were modified with uncertainty-aware loss; the corresponding results are shown in the "Uncertainty-aware Model" columns. Results marked by $*$ are quoted from \cite{luo2022self}. The rest are from our experiments. Results marked by $\dagger$ are from our selected SimCLRv2 approach. The original references for the model architectures are shown next to each model.}
    \vspace{2mm}
    \label{table: pcam2}    
    \begin{tabular}{l  l  l ccc ccc}

        \toprule
          & &  &
            \multicolumn{3}{c}{Regular Model}  &
            \multicolumn{3}{c}{Uncertainty-aware Model} \\
            \textbf{Labels}&\textbf{Training}&\textbf{Model} & Acc (\%) & F1 (\%) & AUC (\%) & Acc (\%) & F1 (\%) & AUC (\%) \\
             
        \midrule
        &Indomain&TransPath$^*$ \cite{wang2021transpath}   & 81.20 & 81.00 & 91.70 & - & - & -\\
        &Indomain&Mocov3$^*$ \cite{chen2021mocov3}  & 86.30 & 86.20 & 95.00 & - & - & -\\
        &Indomain&DINO$^*$ \cite{caron2021emerging} & 85.80 & 85.60 & 95.70 & - & - & -\\
        100\%&Indomain&SD-MAE$^*$ \cite{luo2022self} & 88.20 & 87.80 & 96.20 & - & - & -\\
        &Indomain&MAE \cite{mae}  & 88.41 & 86.23 & 95.81 &  - & - & -\\
        &Indomain&SimCLRv1 \cite{chen2020simple}  & 83.21 & 84.40 & 88.67 & - & - & -\\
        &Indomain&\textbf{SimCLRv2}$^{\dagger}$ \cite{NEURIPS2020_fcbc95cc}   
 & 90.57 & 90.20 & 96.47 &  90.29 & 89.95& 96.49\\
        &Outdomain&\textbf{SimCLRv2}$^{\dagger}$ \cite{NEURIPS2020_fcbc95cc}  & 89.30& 88.97 & 96.58 &  \textbf{91.30} & \textbf{91.09} & \textbf{96.83} \\
        \midrule
        10\%  &Indomain&\textbf{SimCLRv2}$^{\dagger}$ \cite{NEURIPS2020_fcbc95cc}   &  89.73 & 89.07 & 96.19  & 88.27& 88.94 & 94.69 \\
        &Outdomain&\textbf{SimCLRv2}$^{\dagger}$ \cite{NEURIPS2020_fcbc95cc}  & 89.60 & 88.84& 96.73 &  \textbf{90.41} & \textbf{89.97} & \textbf{96.87} \\
        \midrule
        &Indomain&MAE \cite{mae}  & 86.10 & 94.45 & 95.81 &  85.81 & 86.10 & 94.45\\
        1\%&Indomain&SimCLRv1 \cite{chen2020simple}  & 88.67 & 81.52 & 83.45 & 87.77 & 88.67 & 81.52 \\
        &Indomain&\textbf{SimCLRv2}$^{\dagger}$ \cite{NEURIPS2020_fcbc95cc}    &  \textbf{90.27}& \textbf{89.99} & \textbf{95.34}  & 88.96& 88.54 & 94.24 \\
        &Outdomain&\textbf{SimCLRv2}$^{\dagger}$ \cite{NEURIPS2020_fcbc95cc}   & 89.21 & 88.88 & 95.57 &  87.43 & 86.96& 92.33\\
        \bottomrule
    \end{tabular}
    }
\end{table*}



\begin{table*}[]
    \centering
    \small{
    \caption{\small Multi-class classification results for NCT100k dataset for an ensemble of models under 100\%, 10\% and 1\% labels settings. For selected cases, MAE, SimCLRv1, and SimCLRv2 models were modified with uncertainty-aware loss; the corresponding results are shown in the "Uncertainty-aware Model" columns. Results marked by $*$ are quoted from \cite{luo2022self}; Results marked by $**$ are quoted from \cite{jin2022histossl}. The rest are from our experiments. Results marked by $\dagger$ are from our selected SimCLRv2 approach. The original references for the model architectures are shown next to each model.}
    \vspace{2mm}
    \label{table: nct2}    
    \begin{tabular}{l  l  l  cc cc}

        \toprule
          & &  &
            \multicolumn{2}{c}{Regular Model}  &
            \multicolumn{2}{c}{Uncertainty-aware Model} \\
            \textbf{Labels}&\textbf{Training}&\textbf{Model} & Acc (\%) & F1 (\%) & Acc (\%) & F1 (\%)  \\
             
        \midrule
        &Indomain&TransPath$^*$ \cite{wang2021transpath}  & 92.80 & 89.90  & - & - \\
        &Indomain&Mocov3$^*$ \cite{chen2021mocov3}  & 94.40 & 92.60  & - & -\\
        &Indomain&DINO$^*$ \cite{caron2021emerging} & 94.40 & 91.60 & - & - \\
        & Indomain - & BYOL$^{**}$ \cite{grill2020bootstrap}  & 93.93 & -   & - & -\\
        & Indomain &HistoSSL-Res$^{**}$ \cite{jin2022histossl}  & 96.55 & - & - & - \\
        100\%& Indomain &HistoSSL-ViT$^{**}$ \cite{jin2022histossl} & 96.18 & - &  - & - \\  
        &Indomain&SD-MAE$^*$ \cite{luo2022self} & 95.30 & 93.50  & - & - \\
        &Indomain&MAE \cite{mae}  & 94.70 & 94.20   & - & -\\
        &Indomain&SimCLRv1 \cite{chen2020simple}  & 92.10 & 92.20 & -   & - \\
        &Indomain&\textbf{SimCLRv2}$^{\dagger}$ \cite{NEURIPS2020_fcbc95cc}  & 96.28  & 96.25  & 96.44 & 96.39\\
        &Outdomain&\textbf{SimCLRv2}$^{\dagger}$ \cite{NEURIPS2020_fcbc95cc}   & \textbf{96.85}  & \textbf{96.82}   & 95.88 & 95.82\\
        \midrule
        10\%&Indomain&\textbf{SimCLRv2}$^{\dagger}$ \cite{NEURIPS2020_fcbc95cc}  
 & \textbf{96.28}  & \textbf{96.25} & 95.82 & 95.73 \\
        &Outdomain&\textbf{SimCLRv2}$^{\dagger}$ \cite{NEURIPS2020_fcbc95cc}  & 94.62  & 94.56  & 94.98 & 94.87\\
        \midrule
        &Indomain&MAE \cite{mae}   & 93.40 & 92.68 & - & -\\
        1\%&Indomain&\textbf{SimCLRv2}$^{\dagger}$ \cite{NEURIPS2020_fcbc95cc}   & 94.27  & 94.12& 91.70 & 91.65 \\
        &Outdomain&\textbf{SimCLRv2}$^{\dagger}$ \cite{NEURIPS2020_fcbc95cc}  
 & \textbf{94.34}  & \textbf{94.23} & 92.34 & 92.85 \\
        
        \bottomrule
    \end{tabular}
    }
\end{table*}

\begin{figure*}[htbp]
    \centering
    \includegraphics[width=\linewidth]{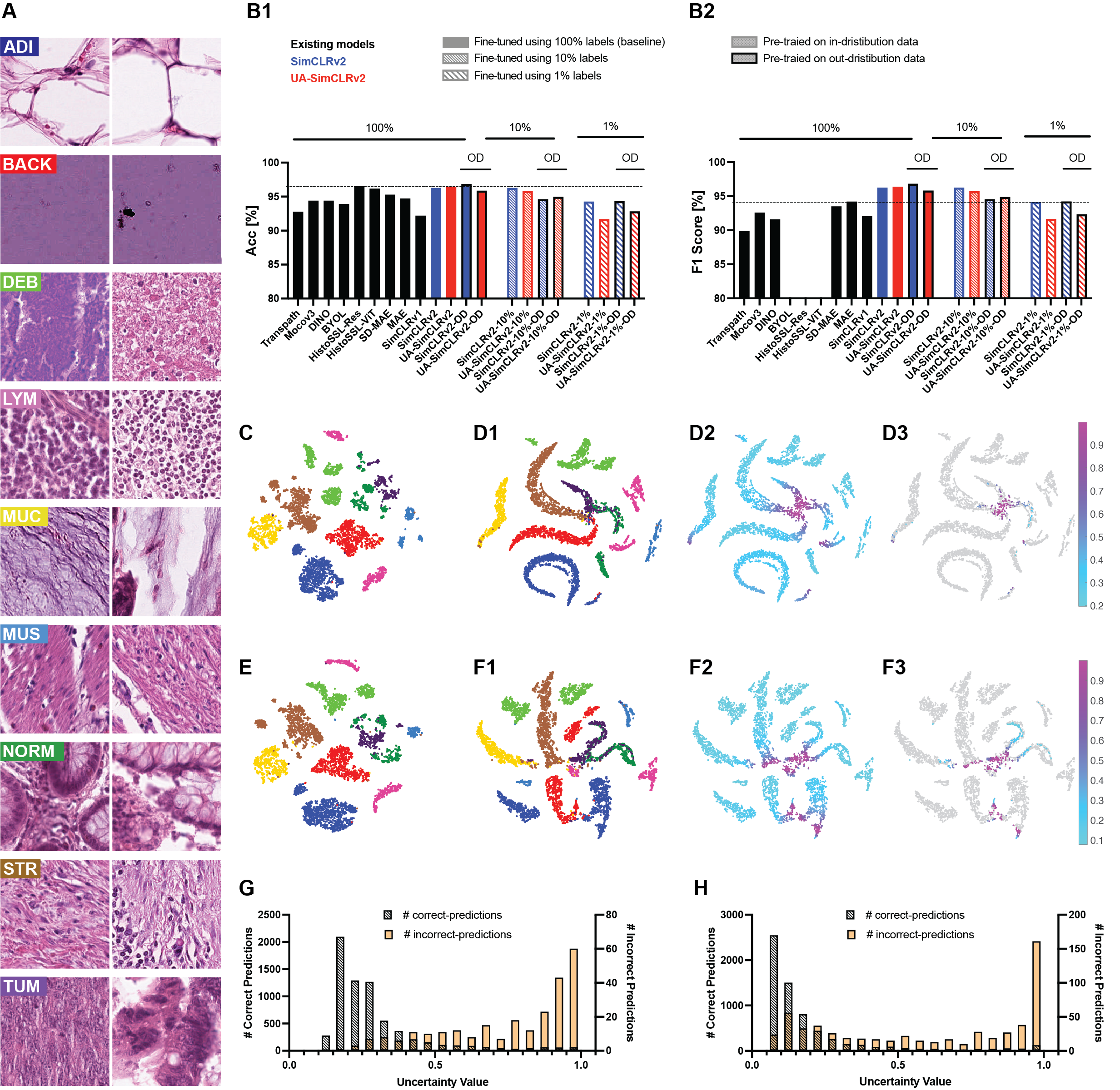}
    \vspace{-6mm}
    \caption{\small Multi-class classification results for NCT100k dataset for an ensemble of models. (A) Representative images from nine classes in the dataset. (B1) Classification accuracy for competing models (shown in black) vs. the proposed models: SimCLRv2 and UA-SimCLRv2 (i.e. Uncertainty Aware SimCLRv2). The proposed models were first trained on 100\% of annotations to generate the baselines. Models were pre-trained using in-distribution data (i.e. training data from NCT100k) as well as out-domain data (i.e. training data from PCam). Then the same experiments were repeated with 10\% of annotations, and 1\% of annotations. (B2) F1-scores for the same experiments in ‘A’. (C) T-SNE plot for SimCLRv2 trained in distribution with 100\% annotations. (D1) T-SNE plot for UA-SimCLRv2 trained in distribution with 100\% annotations. Note that there are four clusters that were hard for the model to separate. (D2) T-SNE plot color coded with the uncertainty values. Note that mixed cluster regions show high uncertainty. (D3) T-SNE plot where only the Incorrect predictions are color coded. Note that most incorrect predictions show high uncertainty. (E, F1, F2, F3) Corresponding versions of ‘C, D1, D2, D3’ with 1\% of annotations. Note that in ‘F3’ there are more incorrect predictions with low uncertainty values than in ‘D3’. (G) Histogram of uncertainty values with correct vs. incorrect predictions for 100\% annotations (H) Same as in ‘G’ with 1\% annotations. Note that incorrect prediction histograms correspond to ‘D3’ and ‘F3’. }
    \label{fig:fig3}
\end{figure*}

\vspace{-3mm}
\noindent \textbf{Multi class classification on NCT100k.} The NCT100k dataset \cite{kather_jakob_nikolas_2018_1214456} comprised 100,000 non-overlapping image patches extracted from hematoxylin and eosin (H\&E) stained histological images of human colorectal cancer and normal tissues. Each patch was 224x224 pixels in size, and annotated with one of nine classes based on the type of tissue: Adipose (ADI), background (BACK), debris (DEB), lymphocytes (LYM), mucus (MUC), smooth muscle (MUS), normal colon mucosa (NORM), cancer-associated stroma (STR), and colorectal adenocarcinoma epithelium (TUM). According to NCT100k dataset's guidelines, we trained our models with NCT100k and tested with CRC-VAL-HE-7K (CRC7k) dataset \cite{kather_jakob_nikolas_2018_1214456}, which consisted of samples from the same data distribution. 

Fig.~\ref{fig:fig3} and Table~\ref{table: nct2} show multi-class classification results for the NCT100k dataset. Similar to the binary case, we experimented at 100\%, 10\%, and 1\% fine-tuning settings. First, at the 100\% setting our SimCLRv2 and UA-SimCLRv2 performed on par with the SOTA. Interestingly, out-distribution-pre-trained SimCLRv2 was the best-performing model and surpassed the SOTA by a small margin. At the 10\% setting, our models still performed comparable to the 100\% baseline and SOTA. But at the 1\% setting, we observed a clear degradation of performance by a few percentage points.  

We further investigated the model behaviors at 100\% and 1\% settings using t-distributed stochastic neighbor embedding (T-SNE) on the learned feature representations. Fig.~\ref{fig:fig3}.C\&E show the T-SNE maps for SimCLRv2 at 100\% \& 1\% settings respectively. Fig.~\ref{fig:fig3}.D1\&F1 show the same for UA-SimCLRv2. Compared to the 100\% setting, the 1\% setting of UA-SimCLRv2 showed more overlapping clusters. For instance, the NORM class heavily overlapped with the TUM class (see Fig.~\ref{fig:fig3}.F1). Fig.~\ref{fig:fig3}.D2\&F2 show the same T-SNE plots from UA-SimCLRv2 in D1\&F1 but color-coded with the associated uncertainty of the predictions. Interestingly, the overlapping regions showed high uncertainty. In Fig.~\ref{fig:fig3}.D3\&F3 we further color-coded only the incorrect predictions. We observed that in the 100\% setting most incorrect predictions were associated with higher uncertainty (Fig.~\ref{fig:fig3}.D3). But in the 1\% setting some incorrect predictions were associated with lower uncertainty (Fig.~\ref{fig:fig3}.F3). We also plotted the histograms of uncertainty values of correct and incorrect predictions (see Fig.~\ref{fig:fig3}.G\&H). For both 100\% and 1\% settings, correct predictions showed a left-skewed distribution; incorrect predictions showed a right-skewed distribution. Thus uncertainty in predictions allows us to identify the data classes that lead to inaccurate predictions. This insight enabled us to develop a sample selection procedure we call \textit{uncertainty-aware training} (UA-training) to fine-tune UA-SimCLRv2. Uncertainty-aware training deviates from the random selection of 1\% annotations employed while keeping the fine-tuning process unchanged.


\begin{figure*}[htbp]
    \centering
    \includegraphics[width=0.9\linewidth]{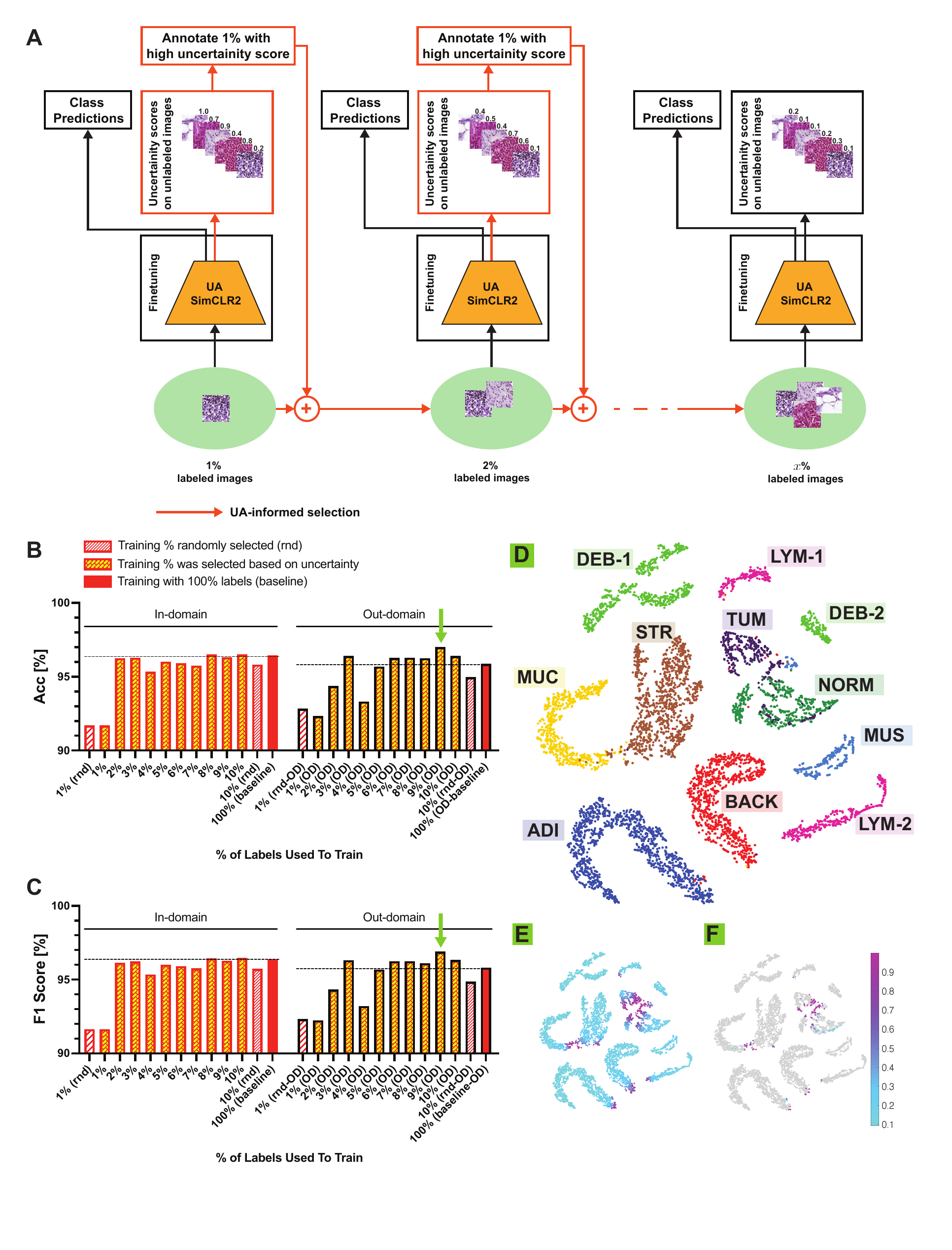}
    \vspace{-11mm}
    \caption{\small (A) Uncertainty-aware fine-tuning of the UA-SimCLRv2 model for NCT100k. UA-SimCLRv2 was pretrained using in-distribution data (i.e. training data from NCT100k) as well as out-domain data (i.e. training data from PCam). Then the pretrained models were finetuned using a randomly selected 1\% of annotations. Next, the uncertainty values were calculated using the 1\%-finetuned model for all remaining training data and another 1\% of labels with high uncertainty score were annotated to have 2\% of annotations for further fine-tuning. This procedure was repeated until the 10\% of annotations were used for fine-tuning. (B) Classification accuracy for Uncertainty-aware fine-tuning vs. randomly selected fine-tuning. (C) F1 score for Uncertainty-aware fine-tuning vs. randomly selected fine-tuning. (D) T-SNE map of the learned features from the final distilled model for the best performing case (i.e. out-domain pretrained model with Uncertainty-aware fine-tuning up to 9\% annotations – also marked using the green arrow in ‘B’). (E) The same T-SNE map in ‘D’ color coded using the uncertainty score. (F) The same T-SNE map in ‘E’ with only wrong classifications color coded. }
    \label{fig:fig4}
\end{figure*}

\subsection{Uncertainty-aware Fine-tuning}
\label{sec:UA-training}
\vspace{-3mm}
\begin{table}[]
    \centering
    \small{
    \caption{\small Results from uncertainty-informed training of UA-SimCLRv2 on the NCT100K dataset.}
    \vspace{2mm}
    \label{table: nct2}    
    \begin{tabular}{l cc cc}

        \toprule
           & \multicolumn{2}{c}{In-domain} &\multicolumn{2}{c}{Out-domain}  \\
            \textbf{Labels}&Acc (\%)& F1 (\%) & Acc (\%) & F1 (\%)  \\
             
        \midrule
        \textbf{1\%}  & 91.70 & 91.65  & 92.34 & 92.25 \\
        \textbf{2\%}  & 96.26 & 96.15  & 94.38 & 94.34\\
        \textbf{3\%} & 96.29 & 96.23 & 96.41 & 96.31 \\
        \textbf{4\%}  & 95.35 & 95.35  & 93.33 & 93.21\\
        \textbf{5\%} & 96.03 & 96.02 & 95.69 & 95.68 \\
        \textbf{6\%}  & 95.93 & 95.91 &  96.28 & 96.25 \\  
        \textbf{7\%}  & 95.76 & 95.76  & 96.29 & 96.25 \\
        \textbf{8\%}  & 96.50 & 96.45   & 96.25 & 96.12\\
        \textbf{9\%} & 96.32 & 96.28 & 97.01   & 96.90 \\
        \textbf{10\%} &  96.51 & 96.49  & 96.40 & 96.33 \\
        
        \bottomrule
    \end{tabular}
    }
\end{table}

The proposed uncertainty-aware training approach is presented in Fig.~\ref{fig:fig4}.A. We started fine-tuning using 1\% randomly selected labeled patches (from the training set of the target task). The resulting model was then inferred on the remaining training set, and uncertainty scores were computed. We then included annotations of the top 1\% patches with the highest uncertainty scores. The annotated patches, totaling 2\% of the training set, were then used for the subsequent fine-tuning step. This iterative process emulates a scenario where only highly uncertain patches are selectively labeled by an expert pathologist. The process was repeated until 10\% of the training set was annotated for fine-tuning. At each step, the fine-tuned UA SimCLRv2 model was saved and evaluated on the CRC7k dataset (i.e. the test set for multi-class classification) to generate test performance metrics.

Fig.~\ref{fig:fig4} and Table~\ref{table: nct2} demonstrate the accuracy and F1 score for uncertainty-aware training. In the in-domain training settings, the accuracy and F1-scores immediately reached just below the 100\%-setting baseline with only 2\% labels; but the performance saturated afterwards. Nevertheless, in both in- and out-domain pre-training settings, UA-training outperformed fine tuning with random selections (compare 2-10\% bars with their corresponding 10\% rnd bars in Fig.~\ref{fig:fig4} [A-B]). The best-performing case was achieved in the out-domain pre-training setting (i.e. UA SimCLRv2 pre-trained on PCam) at 9\% of the labels. Interestingly, this model outperformed both the SOTA, i.e. HistoSSL-Res~\cite{jin2022histossl}, and the 100\% baseline model.  

These results establish the proposed UA-SimCLRv2 as the SOTA patch classifier on NCT100k benchmarks. We next present how the SimCLRv2 and UA-SimCLRv2 models trained on patch-level data can be used as encoders in slide-level classification.


\subsection{Whole Slide Image (WSI) classification using Multiple Instance Learning (MIL)}
\vspace{-3mm}
\begin{table*}[]
\centering
\small{
\caption{\small Whole slide classification results for CAMELYON16 dataset. The column "ImgNet weights" shows the models that used ResNet50 encoders trained on ImageNet in fully supervised settings. The column "Finetuned" shows the percentage of data from the same pre-trained dataset used to finetune our encoders. Results marked by $*$ are quoted from \cite{zhang2022dtfd}}
\vspace{-2mm}
\label{table: dtfd}  
\begin{tabular}{l c l l c c c }
\toprule
Method & ImgNet weights & Pretrained & Finetuned & Acc(\%) & F1(\%) & AUC(\%) \\
\midrule
Mean Pooling$^*$ & \checkmark & - & - & 62.6 & 35.5 & 52.8 \\
Max Pooling$^*$ & \checkmark & - & - & 82.6 & 75.4 & 85.4 \\
RNN-MIL$^*$\cite{campanella2019clinical} & \checkmark & - & - & 84.4 & 79.8 & 87.5 \\
Classic AB-MIL$^*$\cite{abmil} & \checkmark & - & - & 84.5 & 78.0 & 85.4 \\
DS-MIL$^*$\cite{dsmil} & \checkmark & - & - & 85.6 & 81.5 & 89.9 \\
CLAM-SB$^*$ \cite{lu2021data} & \checkmark & - & - & 83.7 & 77.5 & 87.1 \\
CLAM-MB$^*$ \cite{lu2021data} & \checkmark & - & - & 82.3 & 77.4 & 87.8 \\
Trans-MIL$^*$ \cite{shao2021transmil} & \checkmark & - & - & 85.8 & 79.7 & 90.6 \\
DTFD-MIL$^*$ \cite{zhang2022dtfd} & \checkmark & - & - & 89.9 & 86.6 & \textbf{93.3} \\
\midrule
DTFD-MIL + Our encoder & \xmark & PCAM & 100\% & 90.7 &  87.5  &  \textbf{95.5}   \\
DTFD-MIL + Our encoder & \xmark & NCT-CRC & 100\% & 92.2  &  89.6 & 94.0    \\
{DTFD-MIL + Our encoder (UA) }& \xmark & PCAM & 100\% & 87.6 &  82.2  &  93.6   \\
{DTFD-MIL + Our encoder (UA) }& \xmark & NCT-CRC & 100\% & 92.2  &  89.1 & 94.4   \\
\midrule
DTFD-MIL + Our encoder & \xmark & PCAM & 10\% & 86.8   &  81.9 &  89.7  \\  
DTFD-MIL + Our encoder & \xmark & NCT-CRC & 10\% & 89.1  &  86.5 & 93.6    \\
{DTFD-MIL + Our encoder (UA)} & \xmark & PCAM & 10\% & 89.6   &  89.6 &  \textbf{95.9}  \\  
{DTFD-MIL + Our encoder (UA) }& \xmark & NCT-CRC & 10\% & 92.2  &  89.6 & 94.3   \\
\midrule
DTFD-MIL + Our encoder & \xmark & PCAM & 1\% & 85.2   &  80.0  & 88.4   \\
DTFD-MIL + Our encoder & \xmark & NCT-CRC & 1\% & 88.4 &  84.0 & 92.1   \\
{DTFD-MIL + Our encoder (UA)} & \xmark & PCAM & 1\% & 92.2   &  88.8  & \textbf{95.3}  \\
{DTFD-MIL + Our encoder (UA) }& \xmark & NCT-CRC & 1\% & 95.9 &  94.9 & 92.7   \\
\midrule
DTFD-MIL + Our encoder & \xmark & Camelyon-16 & - & 91.5 & 88.8 &  \textbf{95.8} \\

\bottomrule
\end{tabular}}
\end{table*}

In addition to establishing SimCLRv2 as the SOTA patch classifier, we demonstrate its versatility in adapting to whole slide image (WSI) classification at the slide-level. WSI classification is typically performed using multiple instance learning (MIL) under weakly-supervised settings. Here only slide-level annotations are available. We adapted SimCLRv2 and UA-SimCLRv2 for WSI classification on the CAMELYON16 dataset \cite{10.1001/jama.2017.14585}. CAMELYON16 consisted of 400 H\&E WSIs of lymph nodes, with metastatic regions labeled. The dataset consists of two sets: a training set consisting 270 WSIs and a testing set consisting 130 WSIs. 

We adapted DTFD-MIL \cite{zhang2022dtfd}, the SOTA WSI classifier on CAMELYON16, as our WSI classifier (Fig.~\ref{fig:fig1}[D]). DTFD-MIL leverages a ResNet-50 backbone to extract features from patches obtained from WSIs. Features extracted from all the patches (from a particular slide) are then combined using an attention mechanism to predict the slide-level label. In our adaptation, we replaced the ResNet-50 feature extractor (which is pre-trained on ImageNet \cite{DenDon09Imagenet} in a fully supervised setting) with a ResNet-50 that is pre-trained, fine-tuned, and distilled within our proposed SimCLRv2 and UA-SimCLRv2 frameworks using either the CAMELYON16, NCT100k, or PCam datasets. 

\begin{figure*}[htbp]
\vspace{-2mm}
    \centering
    \includegraphics[width=0.9\linewidth]{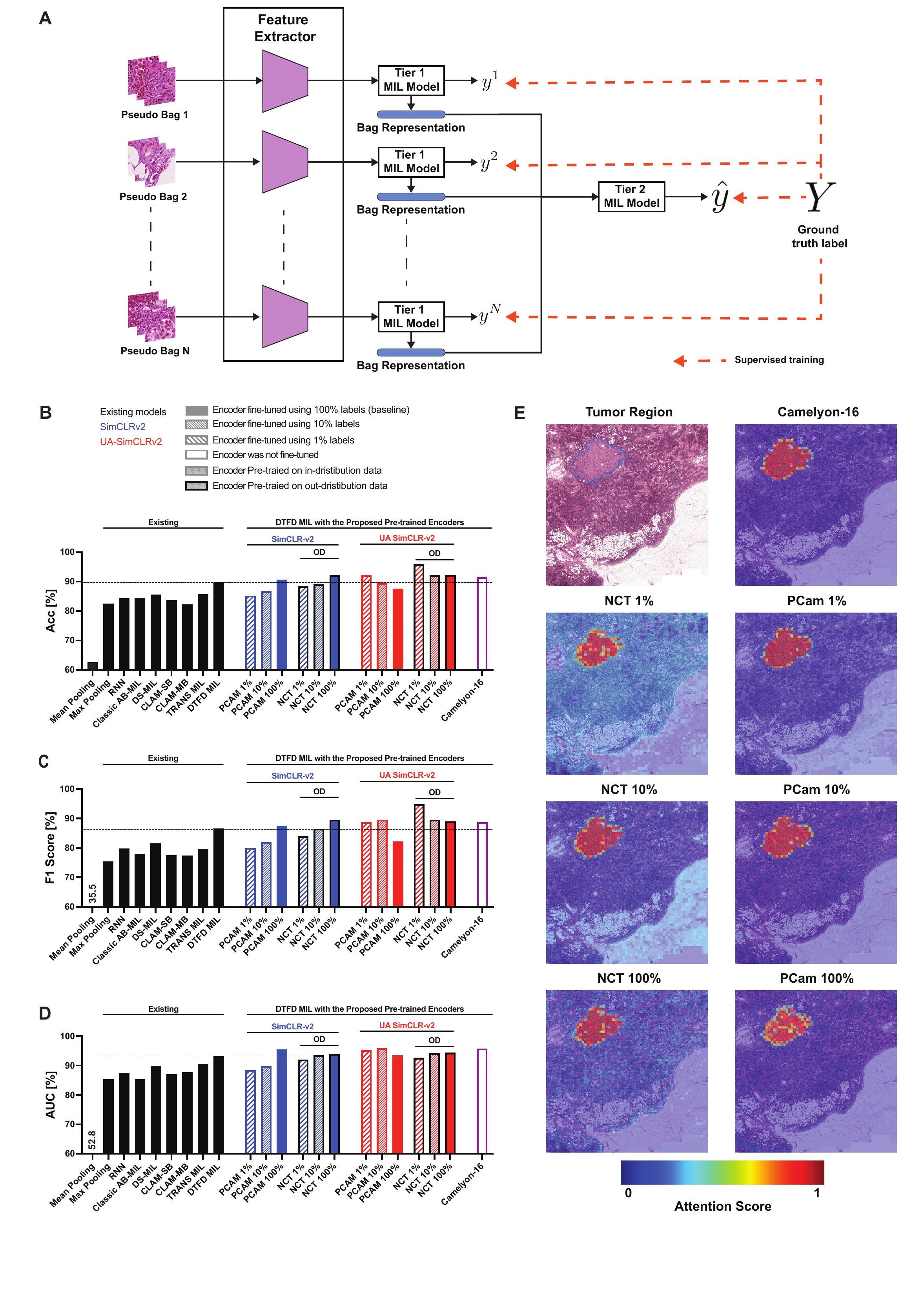}
    \vspace{-2cm}
    \caption{\small Using the proposed models trained on patch level data as encoders for multiple instance learning (MIL). (A) DTFD \cite{zhang2022dtfd} architecture has been used for MIL evaluation. (B) MIL classification accuracy for existing methods  and DTFD using our proposed encoders with multiple combinations of models, \% of annotated patches, datasets. Shown in solid black are the competing methods with DTFD performing the best. We then replaced the encoder of DTFD with our encoders and re-trained the rest of the MIL framework and the results are shown in colored bars. (C) F1 score for the same experiments in ‘B’. (D) Area under the curve (AUC) for the same experiments in ‘B’. Note that, in the spirit of only using slide-level ground truths for training, the camelion-16 pretrained model didn’t use any annotated patch level labels. (E) MIL attention score maps for a representative test slide for different models.}
    \label{fig:fig5}
\end{figure*}

Fig.~\ref{fig:fig5} shows variation of test accuracy, F1 score and AUC scores comparing DTFD-MIL with and without our encoder. The exact values are reported in Table \ref{table: dtfd}. In all 100\%, 10\%, and 1\% settings, some version of the SimCLRv2 and UA-SimCLRv2 models outperformed the SOTA by a few percentage points in all three performance metrics. This result shows the value of introducing a few patch-level annotations to train an encoder for slide-level classification. We further investigated the effect of patch-level contrastive pre-training alone on the slide-level encoder.  In this setting, no patch-level annotations were used to pre-train the encoder. To this end, we pre-trained the SimCLRv2 encoder using patches obtained by splitting the WSIs in CAMELYON16. This model too outperformed the SOTA by a few percentage points, in all three performance metrics (see Camelyon16 case in Fig.~\ref{fig:fig5} and )Table \ref{table: dtfd}).

These results highlight the capability of our proposed approach to achieve accurate and interpretable machine learning with minimal supervision. 


\begin{figure*}[htbp]
    \centering
    \includegraphics[width=\linewidth]{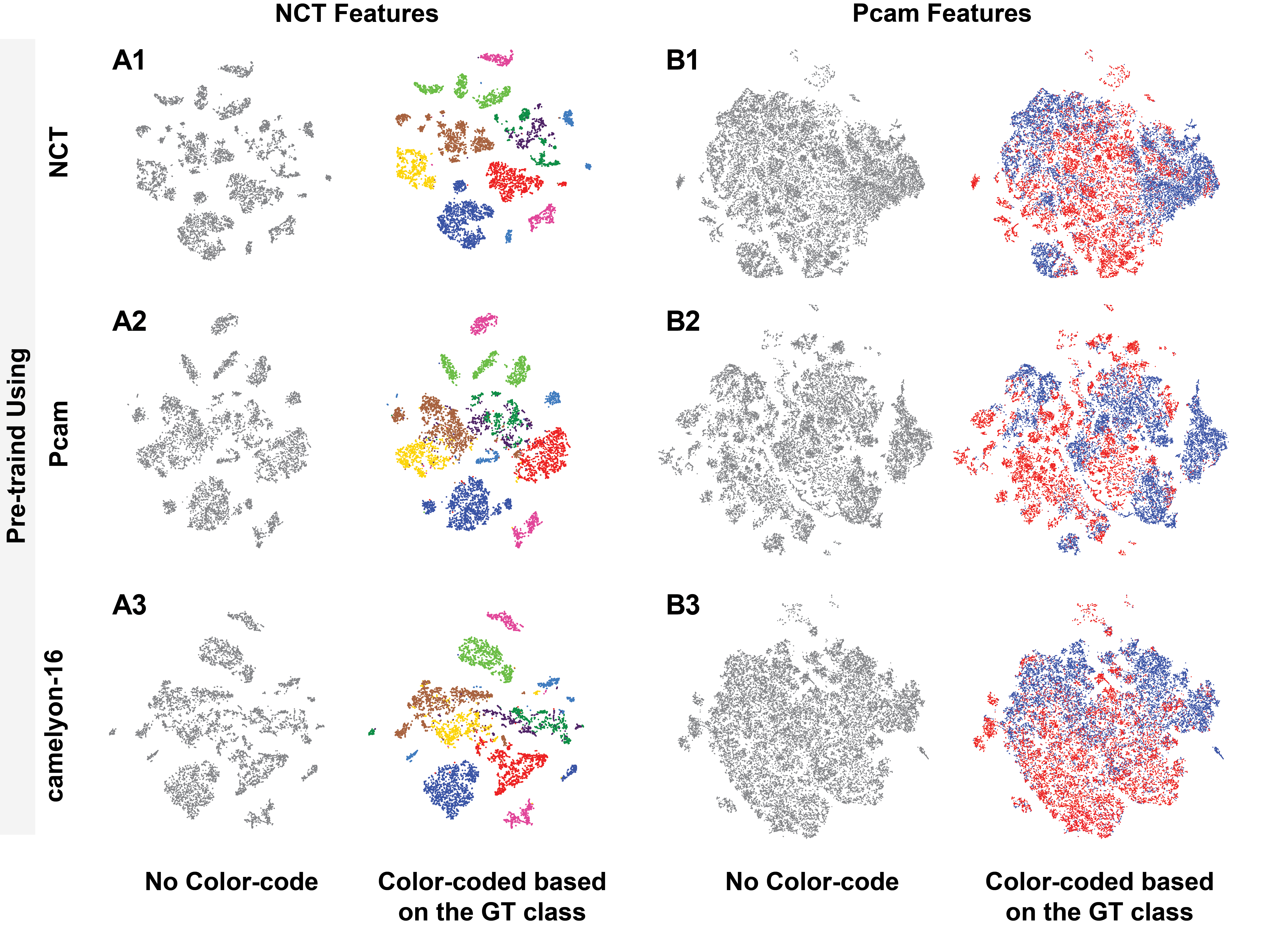}
    \caption{\small Interpreting the pre-training process through T-SNE maps. The first two columns show T-SNE maps for the NCT100k test dataset’s features after pre-training SimCLRv2 using different datasets. Note that the clusters are partially formed for in-domain as well as out-domain pre-training. The remaining two columns show T-SNE maps for the PCam test dataset's features from the same pre-trained SimCLRv2 models. Both non-color-coded and color-coded T-SNE map pairs are provided (based on ground truth labels mentioned in Fig. \ref{fig:fig4}D). (A1) NCT 100K dataset features extracted from SimCLRv2 pre-trained on the NCT100K dataset itself. (A2) NCT 100K dataset features extracted from SimCLRv2 pre-trained on the Pcam dataset. (A3) NCT 100K dataset features extracted from SimCLRv2 pre-trained on the Camelyon-16 dataset. (B1) Pcam dataset features extracted from SimCLRv2 pre-trained on the NCT100K dataset. (B2) Pcam dataset features extracted from SimCLRv2 pre-trained on the Pcam dataset itself. (B3) Pcam dataset features extracted from SimCLRv2 pre-trained on the Camelyon-16 dataset.}
    \label{fig:fig6}
\end{figure*}

\section{Discussion and Conclusion}
\vspace{-1mm}
In summary, based on the seminal SimCLRv2 self-supervised framework \cite{chen2020big}, we introduce an uncertainty-aware contrastive learning method for digital pathology. Through a series of experiments, we showcase the performance of our framework across various histopathology tasks, even when confronted with limited annotated data. Furthermore, we address a critical limitation in histopathology models by incorporating uncertainty awareness, which has significantly enhanced the interpretability of our novel model. This newfound interpretability may empower clinicians to make well-informed decisions, facilitating the seamless integration of AI into the manual clinical decision-making process.

Our findings indicate that vision models with CNN backbones outperform transformer-based models in histopathology visual learning. Specifically, SimCLRv2, equipped with a simple ResNet-50 backbone, surpasses current SOTA models. It has the potential to achieve even better performance with a deeper ResNet backbone. The superiority of SimCLRv2 may be due to contrastive learning. As depicted in Fig.~\ref{fig:fig6}, the feature clusters formed by the SSL pre-trained model reveal its ability to differentiate distinct classes and their corresponding features, even before any annotations are introduced. The learning procedure of SimCLRv2, which effectively utilizes a large unlabelled dataset in two steps of the pipeline, confers the advantage of learning a highly effective encoder for patch-level data.

When a sufficient number of annotations are available for UA training, UA SimCLRv2 tends to outperform SimCLRv2. The transition from the feature space clusters of the SimCLRv2 model to those of UA SimCLRv2 results in more well-defined and tightly shaped clusters, indicating an improved ability to classify accurately (see Fig.~\ref{fig:fig3}). Moreover, the alignment of high uncertainty with incorrect predictions demonstrates the model's capability to identify challenging cases and exhibit lower confidence in predicting them.

Our results also suggest that when enough labels are available for fine-tuning, pre-training on the out-distribution data results in higher performance. The superiority of this out-domain trained model can be attributed to the advantage gained from a large number of classes available for contrastive pre-training. With an increased diversity of classes, the encoder can effectively compare and cluster features, leading to a better understanding of distinctive tissue characteristics and the establishment of clear boundaries within the feature space. Last, the seamless integration of our models into the WSI classification demonstrates the versatility and adaptability of our approach.

However, in scenarios with fewer annotations, such as our 1\% setting, UA-SimCLRv2 exhibits significant underperformance, often providing incorrect predictions with high confidence, particularly when multiple data classes are present. This poses a substantial risk in the context of medical AI, as the model's confidence does not necessarily reflect the correctness of its decisions. Thus we recommend using approaches such as uncertainty-aware fine-tuning (proposed in the section ~\ref{sec:UA-training}) to mitigate such risks when fine-tuning with few annotations. We also highly recommend rigorous testing with independent test datasets that aren't used during any stage of the training process including the hyper-parameter estimation. 

In conclusion, our extensive patch- and slide-level experiments with contrastive learning and uncertainty quantification set new benchmarks in digital pathology classification. Our results consistently suggest the efficient use of large datasets -despite having no annotations- helps build better models for digital pathology. We believe that our work sets the foundation upon which multiple clinical tasks can use large digital pathology datasets efficiently, accurately, and interpretably.

\subsubsection*{ACKNOWLEDGEMENTS}
\vspace{-1mm}
This work was supported by the Center for Advanced Imaging and the John Harvard Distinguished Science Fellowship Program within the FAS Division of Science of Harvard University.

\vspace{10mm}

\vspace{-2mm}
\section{Methodology}
\label{sec:Methodology}
\vspace{-1mm}
\subsection{UA SimCLRv2}
\label{sec:UQ}
\vspace{-2.5mm}

To address the critical need for interpretability, we introduce UA-SimCLRv2. The primary objective of UA-SimCLRv2 is to enhance the interpretability of the model's predictions in the context of histopathology analysis. This is achieved by incorporating the theory of uncertainty estimation, which serves as the basis for uncertainty awareness in UA SimCLRv2.

In \cite{sensoy2018evidential}, the uncertainty estimation is approached from  Dempster–Shafer theory of evidence (DST) perspective \cite{dempster2008upper} assigning belief masses to subsets of a frame of discernment, which denotes the set of exclusive possible states. Subjective logic (SL) formalizes DST’s notion of belief assignments over a frame of discernment as a Dirichlet distribution. Term evidence is a measure of the amount of support collected from data in favor of a sample to be classified into a certain class. Through model training evidence $e_k$ ($k=1,2...K$)are collected and belief masses $b_k$ ($k=1,2...K$) are assigned to each class based on the evidence collected and the remaining are marked as uncertainty $u$. For $K$ mutually exclusive classes,

\vspace{-3mm}
\begin{equation}\label{eq:uncer plus bm}
    u + \sum_{k=1}^{K}b_k = 1
\end{equation}
Here $u\geq0$ and $b_k\geq0$ and they are calculated by,
\vspace{-3mm}
\begin{equation}\label{eq:uncer and bm}
    b_k=\frac{e_k}{S}  \quad \text{and} \quad  u=\frac{K}{S}, \quad \text{where} \quad S=\sum_{i=1}^{K}e_i + 1
\end{equation}
Observe that when there is no evidence, the belief for each class is zero and the uncertainty is one. A belief mass assignment, i.e., subjective opinion, corresponds to a Dirichlet distribution with parameters $\alpha_k=e_k + 1$. A Dirichlet distribution parameterized over evidence represents the density
of each such probability assignment; hence it models second-order probabilities and uncertainty \cite{josang2016generalising}. It is characterized by $K$ parameters $\alpha=[\alpha_1,\alpha_2, .... , \alpha_K]$ and is given as follows.
\label{eq:dirichlet}
\[
{
    D(p\|\alpha) =
    \begin{cases}
    \frac{1}{B(\alpha)}\prod_{i=1}^{K}p_i^{\alpha_i - 1}, & \text{if } p \in S_K \\
    0, & \text{otherwise}
\end{cases}
}  
\]
Where $S_K = \{p\| \sum_{i=1}^{K}p_i =1 \quad \text{and} \quad 0\leq p_1,...p_k \leq 1 \}$ and $B(\alpha)$  is the $K$-dimensional multinomial beta function \cite{kotz2000continuous}.

Model training follows the classical neural network architecture with a softmax layer replaced with ReLU activation layer to ascertain non-negative output, which is taken as the evidence vector for the predicted Dirichlet distribution. For network parameters $\theta$, let $f(x_i\|\theta)$ be the evidence vector predicted by the network for the classification. Corresponding Dirichlet distribution's parameters $\alpha_i=f(x_i\|\theta) + 1$ are calculated and its mean $(\frac{\alpha_i}{S})$ is considered as the class probabilities. Let $y_i$ be one hot vector encoding the ground-truth class label of a sample $x_i$. Treating $D(p_i\|\alpha_i)$ as a prior on the sum of squares loss $\|y_i - p_i\|_2^2$, obtain the loss function 
\vspace{-3mm}
\begin{align}\label{eq:mseloss}
L_i(\theta)=\int\|y_i - p_i\|_2^2\frac{1}{B(\alpha_i)}\prod_{i=1}^{K}p_{ij}^{\alpha_{ij} - 1}dp_i
\end{align}

By decomposing the first and second moments, minimization of both the prediction error and the variance of the Dirichlet experiment for each sample is achieved by the above loss function. Further some evidence collected might strengthen the belief for multiple classes. To avoid situations where evidence with more ambiguity assigns more belief to incorrect class,  Kullback-Leibler (KL) divergence term is appended to the loss function. Following is the total loss used for UA fine-tuning.

\vspace{-3mm}
\begin{align}\label{eq:loss}
L(\theta)=&\sum_{i=1}^NL_i(\theta) \notag \\ &+\lambda_t\sum_{i=1}^NKL[D(p_i\|\Tilde{\alpha_i})\|D(p_i\|<1,...,1>)]
\end{align}
where $\lambda_t=\min(1.0,t/10) \in [0,1]$ is the annealing coefficient, $t$ is the index of the current training epoch, $D(p_i\|<1,...,1>)$ is the uniform Dirichlet distribution, and $\Tilde{\alpha_i}=y_i+(1-y_i)*\alpha_i$ is the Dirichlet parameters after removal of the non-misleading evidence from predicted parameters $\alpha_i$ for sample $i$. The KL divergence term in the loss can be calculated as

\begin{multline*}\label{eq:KL}
KL[D(p_i\|\Tilde{\alpha_i})||D(p_i\|1)] \\
= \log\bigg(\frac{\Gamma(\sum_{k=1}^K\Tilde{\alpha_{ik}})}{\Gamma(K)\prod_{k=1}^K\Gamma(\Tilde{\alpha_{ik}})}\bigg) \\
+ \sum_{k=1}^K(\Tilde{\alpha_{ik}}-1)\bigg[\psi(\Tilde{\alpha_{ik}}) - \psi\bigg(\sum_{j=1}^K\Tilde{\alpha_{ij}}\bigg)\bigg] 
\end{multline*}
where 1 represents the parameter vector of $K$ ones, $\Gamma$ is the gamma function, and $\psi$ is the digamma function. By gradually increasing the effect of the KL divergence in the loss through the annealing coefficient, the neural network is allowed to explore the parameter space and avoid premature convergence to the uniform distribution for the misclassified samples, which may be correctly classified in future epochs.
\vspace{-3mm}

\subsection{DTFD-MIL framework for MIL}
\label{sec:dtfd}
\vspace{-2.5mm}
DTFD-MIL \cite{zhang2022dtfd} uses a pseudo bag concept to virtually increase the number of bags and uses a doublet-tier approach for WSI classification. One WSI bag is randomly divided into multiple pseudo bags with relatively less number of patches. A pseudo bag is given the label of the parent bag. DTFD-MIL is applied on top of the pseudo bags to predict a bag label. It uses the commonly used attention based MIL approach in each tier. First, patch features are extracted from each pseudo bag using a ResNet backbone. These features are forwarded to the attention based tier-1 model which computes attention scores and instance probabilities for each patch. Tier-1 model aggregates patch features into an embedding that represents the pseudo bag. A feature distillation is performed on top of patch embeddings, using tier-1 instance probabilities, to extract a distilled feature vector. Distilled feature vectors from all pseudo bags are forwarded to the tier-2 attention based model, which aggregates them using attention to learn the final bag embedding for the parent bag. Bag labels from all of the tier-1 models and tier-2 model is compared with ground truth parent bag label to drive the cross entropy loss for training. 



{
\bibliographystyle{IEEEtran}
\bibliography{main}}

\renewcommand{\thefigure}{S\arabic{figure}}
\setcounter{figure}{0}    

\begin{figure*}[h]
    \centering
    \includegraphics[width=0.8\linewidth]{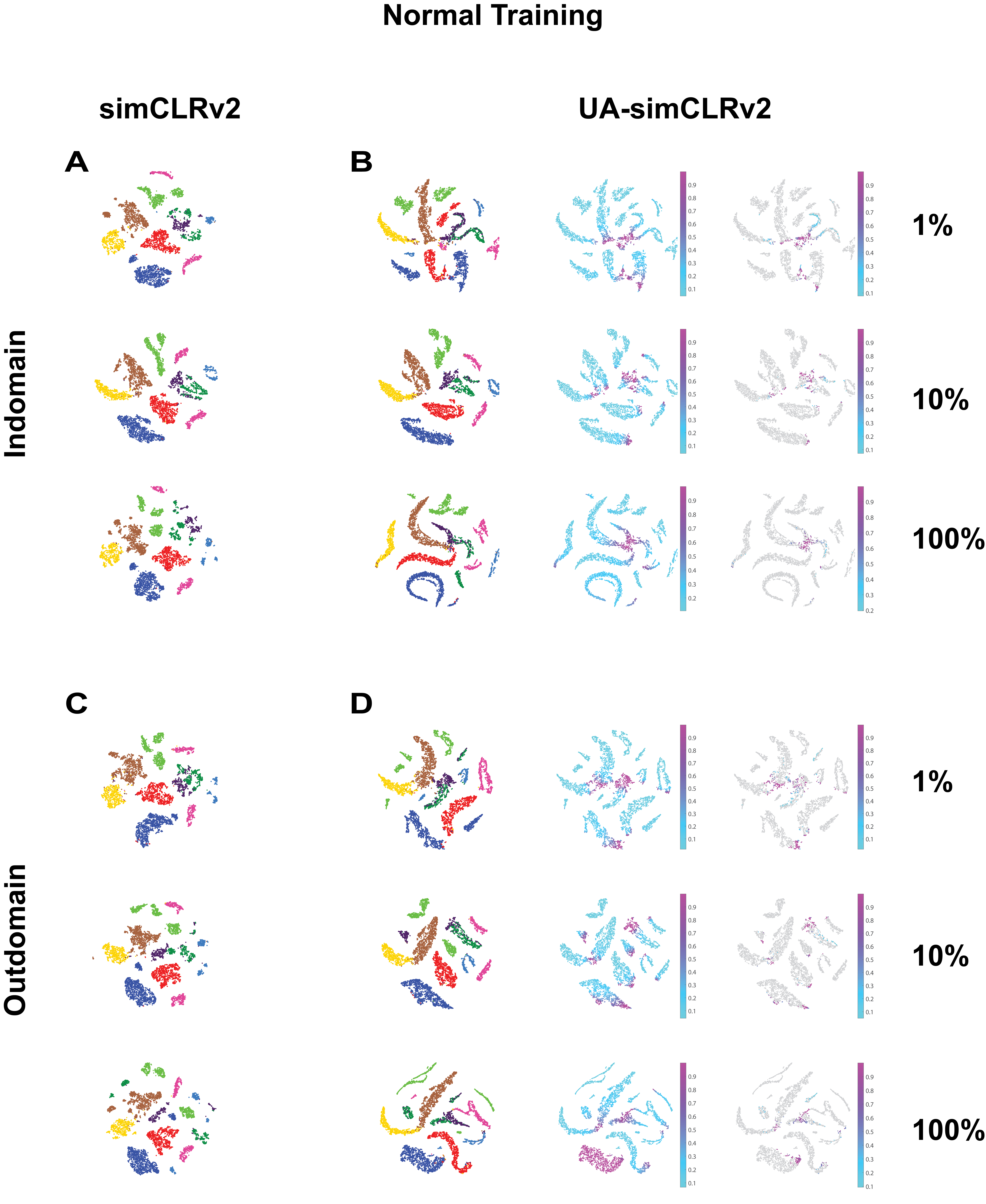}
    \caption{\small {Variation of NCT100K dataset feature space learnt by SimCLRv2 and UA-SimCLRv2 based on a percentage of annotations and the pre-training distribution. In the "Indomain" setting, the model undergoes pre-training and fine-tuning on the NCT100k. In the "Outdomain" setting, the model undergoes pre-training on the PCam dataset and fine-tuning on the NCT100K. The T-SNE maps on the first two columns are colour-coded based on NCT100K classes as shown in Fig.\ref{fig:fig4}D. The third column has T-SNE maps color-coded based on their uncertainty scores and the fourth column T-SNE maps has only the misclassified points highlighted.  (A) and (B) showcase T-SNE maps for features extracted from SimCLRv2 and UA-SimCLRv2 models respectively in the "Indomain" setting. Similarly, (C) and (D) showcase T-SNE maps for features extracted from SimCLRv2 and UA-SimCLRv2 models respectively in the "Outdomain" setting.} }
    \label{fig:fig3-1}
\end{figure*}

\begin{figure*}[h]
    \centering
    \includegraphics[width=0.5\linewidth]{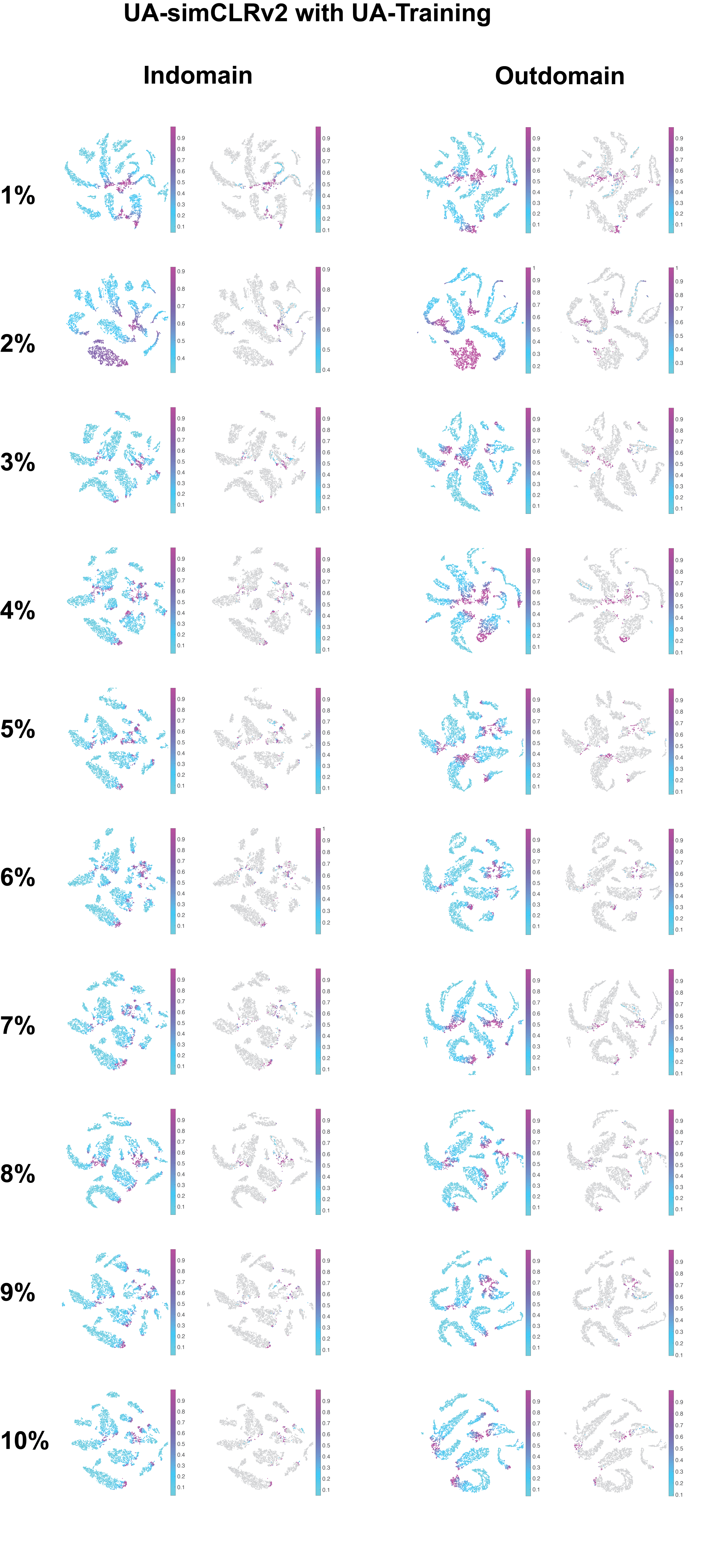}
    \caption{\small {Variation of NCT100K dataset feature space learnt by UA-SimCLRv2 in selective labeling approach based on a percentage of annotations and the pre-training distribution. In the "Indomain" setting, the UA-SimCLRv2 model undergoes pre-training, fine-tuning and UA-Training on the NCT100k dataset. In the "Outdomain" setting, the UA-SimCLRv2 model undergoes pre-training on the PCam dataset and fine-tuning and UA-Training on the NCT100K dataset. We present paired T-SNE maps for each scenario. The first map is color-coded by the uncertainty score whereas the second map in each pair highlights only the misclassified instances.}}
    \label{fig:fig4-2}
\end{figure*}



\end{document}